\documentclass[conference]{IEEEtran}
\IEEEoverridecommandlockouts
\usepackage{amsmath,amsfonts}
\usepackage{array}
\usepackage{textcomp}
\usepackage{stfloats}
\usepackage{url}
\usepackage{amsmath, amssymb, amsthm}
\usepackage{verbatim}
\usepackage{graphicx}
\usepackage{amssymb}
\usepackage{hyperref}
\usepackage{graphicx}
\usepackage{amsmath}
\usepackage{longtable}
\usepackage{algorithm} 
\usepackage{algpseudocode} 
\usepackage{mathrsfs}
\usepackage{subcaption}
\usepackage{mathtools}
\usepackage{pifont}
\usepackage{color}
\usepackage{lineno}
\usepackage{graphicx}  % For including graphics
\usepackage{makecell} 
\usepackage{pdflscape}
\usepackage{adjustbox}
\usepackage[utf8]{inputenc}
\usepackage{tabularx}
\usepackage{blindtext}
\usepackage{longtable}
\usepackage{lscape}
\usepackage{amsthm}
\usepackage{graphicx}
\usepackage{subcaption} % Required for subfigures

\usepackage{setspace}
\usepackage{notoccite} %citation number ordering
\usepackage{lscape} %landscape table
\usepackage{mwe}%%%%%%%%%%%%%%%
\usepackage{booktabs}
\usepackage{amsthm}

\theoremstyle{definition}

\theoremstyle{definition}

\newcommand{\RNum}[1]{\lowercase\expandafter{\romannumeral #1\relax}}
\newcommand{\RNumU}[1]{\uppercase\expandafter{\romannumeral #1\relax}}
\usepackage[numbers]{natbib}

\def\BibTeX{{\rm B\kern-.05em{\sc i\kern-.025em b}\kern-.08em
    T\kern-.1667em\lower.7ex\hbox{E}\kern-.125emX}}
\begin{document}

\title{\title{GBFRVFL: Granular-Ball Computing-Based Fuzzy Random Vector Functional Link Network}
% {\footnotesize \textsuperscript{*}Note: Sub-titles are not captured for https://ieeexplore.ieee.org  and
}

\author{
\IEEEauthorblockN{A. Quadir}
\IEEEauthorblockA{
\textit{Department of Mathematics} \\
\textit{Indian Institute of Technology Indore}\\
mscphd2207141002@iiti.ac.in}
\and
\IEEEauthorblockN{A. Rahaman}
\IEEEauthorblockA{
\textit{Department of Mathematics} \\
\textit{Indian Institute of Technology Indore}\\
phd2401141001@iiti.ac.in}
\and
\IEEEauthorblockN{P. N. Suganthan}
\IEEEauthorblockA{
\textit{Department of Computer Science}\\ 
\textit{and Engineering, College of Engineering} \\
\textit{Qatar University, Qatar}\\
p.n.suganthan@qu.edu.qa}
\and
\IEEEauthorblockN{M. Tanveer\textsuperscript{*}\thanks{\textsuperscript{*}Corresponding author}}
\IEEEauthorblockA{
\textit{Department of Mathematics} \\
\textit{Indian Institute of Technology Indore}\\
mtanveer@iiti.ac.in}
}

\maketitle
\begin{abstract}
In practical machine learning tasks, data are often contaminated with noise, outliers, and class imbalance, which can degrade the performance of conventional models. While random vector functional link (RVFL) networks offer fast training and strong generalization, they do not explicitly handle uncertainty or exploit local data structure. To address these limitations, we propose a fuzzy granular-ball random vector functional link (GBFRVFL) framework that leverages granular-ball computing to abstract raw samples into adaptive granular balls. Within this framework, we introduce two membership assignment schemes: (i) F-GBRVFL, which incorporates fuzzy membership to quantify the reliability of each granular ball, and (ii) SDAP-GBRVFL, which we propose, incorporates a novel statistical density-adaptive pythagorean membership (SDAPM) scheme that dynamically adjusts membership and non-membership values based on class variance, local sparsity, and granular-ball compactness. These schemes enhance robustness to noise, outliers, class imbalance, and uncertainty in granular-ball distributions, while retaining the computational efficiency of RVFL networks. Extensive experiments on 37 benchmark UCI and KEEL datasets under both clean and noisy conditions demonstrate that the proposed models consistently outperform baseline models, achieving superior accuracy and stability. The results validate the effectiveness of integrating granular-ball computing with adaptive membership schemes for reliable, scalable, and noise-tolerant learning. The code and supplementary material of the paper can be accessed using the following link: \url{https://github.com/mtanveer1/GBFRVFL}.
\end{abstract}
\section{Introduction}
\IEEEPARstart{A}{r}tificial neural networks (ANNs), including single hidden layer feedforward neural (SLFN) architectures, have been widely applied to classification and regression tasks \cite{hornik1989multilayer, leshno1993multilayer}, owing to their strong universal function approximation properties \cite{igelnik1995stochastic}. A traditional approach for training SLFNs is the backpropagation (BP) algorithm \cite{lecun1989handwritten}; however, it often encounters issues such as slow convergence, susceptibility to local minima, and sensitivity to the choice of learning rate \cite{quadir2026garfln, quadir2026hypergraph}. To address these limitations, several learning models based on randomization have been developed, including the Schmidt network \cite{schmidt1992feed}, random vector functional link (RVFL) neural networks \cite{pao1994learning}, extreme learning machines (ELM) \cite{huang2006extreme}, and radial basis function networks (RBFN) \cite{park1993approximation}. Among these, the RVFL network has a simple architecture and efficient performance. The RVFL training procedure is carried out in two distinct phases. Initially, the weights and biases connecting the input layer to the hidden layer are randomly assigned within a predefined range \cite{zhang2016comprehensive} and remain unchanged during training. Subsequently, the output-layer parameters are computed analytically using a closed-form solution, enabling efficient and stable learning \cite{malik2023random}. Due to its fast training speed and strong generalization capability, the RVFL network has been successfully applied in a wide range of domains, including time-series prediction \cite{gao2021walk}, hyperspectral image classification \cite{quadir2025randomized}, prediction of DNA-binding proteins \cite{quadir2024multiview}, and nonlinear system identification \cite{chakravorti2020non}. 

Despite their efficiency, RVFL models and their variants often degrade in performance on noisy or imperfect datasets because they assign equal importance to all training samples, making them sensitive to noise, outliers, and class imbalance. Fuzzy theory has been widely used to mitigate such issues \cite{rezvani2019intuitionistic, quadir2024intuitionistic}. Granular computing (GC) provides a framework for handling uncertainty and incomplete data through information granules, as introduced by \citet{zadeh1996fuzzy}. Significant advances have been made by integrating GC with machine learning techniques such as rough sets, fuzzy sets, three-way decision models, and formal concept analysis \cite{ding2022novel, zhang2021granular}. Inspired by the global precedence characteristic of human cognition \cite{chen1982topological}, multigranularity frameworks have been developed to enhance uncertainty handling \cite{wang2017dgcc, quadir2024enhanced}. Recent works include the granular-ball twin support vector machine (GBTSVM) \cite{10759815, 10619989} and granular ball RVFL (GB-RVFL) \cite{sajid2025gb}, which use granular balls instead of individual samples. While these approaches improve performance on complex, high-dimensional data, they remain vulnerable to noise and outliers that can distort granular representations and reduce classification reliability.

To address the challenges posed by noise, outliers, and uncertainty within granular balls, we developed a novel membership assignment mechanism, termed statistical density-adaptive pythagorean membership (SDAPM), and incorporated it into the GBRVFL network, resulting in the proposed SDAP-GBRVFL model. In addition, a fuzzy membership scheme is also integrated in the GBRVFL framework to call F-GBRVFL.  Unlike conventional granular-ball classifiers, which often rely on global thresholds or fixed radii and are therefore highly sensitive to extreme samples, the SDAPM scheme adaptively adjusts membership degrees based on local statistical properties, including class-wise variance and neighborhood sparsity. By integrating this adaptive membership into the GBRVFL framework, the model leverages GBRVFL’s simple architecture and fast training while benefiting from the coarse-grained, noise-resilient representation of granular balls.

The paper’s key highlights are as follows:
\begin{enumerate}
    \item We propose a density-adaptive Pythagorean membership scheme that dynamically adjusts membership values using class variance, sparsity, and compactness to handle noise, outliers, and class imbalance.
    \item We propose F-GBRVFL and SDAP-GBRVFL, which use fuzzy and density-adaptive membership schemes to improve robustness to noise, outliers, class imbalance, and uncertain granular-ball distributions.
    \item Using granular balls instead of individual samples reduces computation by inverting only the GB-center matrix, while preserving discriminative information and reducing noise and outlier effects.
    \item F-GBRVFL and SDAP-GBRVFL are evaluated on KEEL and UCI benchmark datasets under clean and noisy labels and compared with state-of-the-art models.
\end{enumerate}

\section{Related Work}
In this section, we provide an overview and discussion of the granular ball computing (GBC) methodology.
\subsection{Granular-Ball Computing}
The core idea of GBC is to represent the original dataset using a collection of granular balls, where each granular ball summarizes a group of samples. Let $\mathcal{M} = \{(x_1,l_1),(x_2,l_2),\ldots,(x_n,l_n)\}$ denote a dataset, where $X = \{x_j \mid x_j \in \mathbb{R}^d,, j=1,2,\ldots,n\}$ represents the feature matrix consisting of $d$ attributes, and $L = \{l_j \mid l_j \in \mathbb{R}, j=1,2,\ldots,n\}$ denotes the associated label set. The entire sample space $\mathcal{D}$ is then partitioned or covered by a collection of granular balls $\mathcal{GB} = \{GB_i \mid i=1,2,\ldots,k\}$. For the $i^{th}$ granular-ball $GB_i$, its center is defined as $c_i = \frac{1}{n_i}\sum_{k=1}^{n_i} x_{ij}$, where $x_{ij}$ denotes the $j^{th}$ sample belonging to $GB_i$, and $n_i$ is the total number of samples contained in this granular-ball. The radius $r_i$ of $GB_i$ can be computed using the average-distance, $r_i = \frac{1}{n_i}\sum_{k=1}^{n_i} |x_{ij} - c_i|.$ To reduce the influence of noisy samples inside a granular ball, a representative label is assigned to each $GB_i$ based on majority voting. Specifically, the label $y_i$ is determined as the class that occurs most frequently among the samples contained in $GB_i$. The purity of the granular ball denoted by $p_i$ is then defined as the ratio between the number of samples in $GB_i$ that share the dominant label $y_i$ and the total number of samples $n_i$ within the granular ball. 

Within the GBC framework, the first step is to construct granular balls that represent the entire dataset.
This is achieved by generating granular balls through the following optimization problem:
\begin{align}
    & \min~\gamma_1 \sum_{GB_i \in \mathcal{GB}} |GB_i| + \gamma_2 k, \nonumber \\
    & \text{s.t} ~ \text{quality}(GB_i) \geq \rho,
\end{align}
where $\gamma_1$ and $\gamma_2$ are weighting parameters, $k$ denotes the total number of granular balls, and $\text{quality}(GB_i)$ measures the fraction of samples sharing the dominant label within the granular ball $GB_i$. 

\section{The Proposed Fuzzy Granular-Ball Random Vector Functional Link Network}
In real-world data, noise, outliers, and class overlap often reduce model reliability. Although RVFL networks are fast and generalize well, they do not explicitly capture uncertainty or local relationships. Granular-ball computing addresses this by grouping similar samples into adaptive granular balls, while fuzzy theory enhances robustness by assigning membership degrees that reflect granule reliability. Building on this, we propose two variants: F-GBRVFL, which uses a fuzzy membership scheme to handle noisy and ambiguous samples, and SDAP-GBRVFL, which incorporates a statistical density-adaptive Pythagorean membership (SDAPM) scheme to dynamically adjust membership and non-membership based on class variance, local sparsity, and granular-ball compactness.

\subsection{Fuzzy Membership}
The membership degree of a fuzzy granular ball reflects its reliability and importance in learning. Two cases are considered for determining $\delta_{GB_i}$. If prior membership information is available, the granular-ball membership is computed as the average of its samples. Since this is usually unavailable, a suitable membership function must be designed. We adopt a simple and effective distance-based strategy, where samples closer to the class center receive higher membership values, while distant samples receive lower values due to greater uncertainty.

Let $x^+$ and $x^-$ denote the mean vectors of the positive and negative classes, respectively. The class radii are defined as:
\begin{align}
    r^+ = \max_{x_i, y_i=+1} |x_i - x^+|, \quad
    r^- = \max_{x_i, y_i=-1} |x_i - x^-|.
\end{align}
Based on these quantities, the fuzzy membership of a sample $x_i$ is computed as:
\begin{align}
    \mu(x_i) =
\begin{cases}
1 - \dfrac{|x_i - x^+|}{r^+ + \varepsilon}, & y_i = +1, \\
1 - \dfrac{|x_i - x^-|}{r^- + \varepsilon}, & y_i = -1,
\end{cases}
\end{align}
where $\varepsilon > 0$ is a small constant introduced to prevent zero membership values. Finally, the membership degree of a fuzzy granular-ball $GB_i$ is obtained by aggregating the memberships of the samples contained within it, defined as:
\begin{align}
    \delta_{GB_i} = \frac{1}{|GB_i|} \sum_{x_j \in GB_i} \mu(x_j).
\end{align}

\subsection{Statistical Density-Adaptive Pythagorean Membership (SDAPM)}
The baseline membership assignment depends strongly on the maximum radius $R_{max}$, making it sensitive to outliers. To address this issue, we propose a scheme based on class-wise variance and local sparsity.
\subsubsection{Statistical Membership Degree}
Instead of a linear distance decay, we utilize a Gaussian Statistical Kernel \cite{quadir2025twin, quadir2025trkm}. We define the class center $C_j$ and the standard deviation of distances $\sigma_j$ for each class $j \in \{+1, -1\}$. The membership degree of a granular-ball $GB_i$ to its assigned class $y_i$ is defined as:
\begin{align}
    \mu_{GBi} = \exp\left( -\frac{\|c_i - C_{y_i}\|^2}{2(\sigma_{y_i} \cdot \lambda_{y_i})^2} \right),
\end{align}
where $\lambda_{y_i} = \sqrt{ \frac{m}{m_{y_i}}}$ represents the imbalance balancer. 
\subsubsection{Non-membership via Local Sparsity Index}
When purity is unavailable, the non-membership degree is defined using a local sparsity index based on the proportion of nearby granular balls with different labels:
\begin{align}
    LSI_i = \frac{1}{k} \sum_{j \in KNN(GB_i)} \mathbb{I}(y_j \neq y_i).
\end{align}
The pythagorean non-membership degree $\nu_{GBi}$ is then derived such that it satisfies the constraint $\mu_{GBi}^2 + \nu_{GBi}^2 \leq 1$:
\begin{align}
    \nu_{GBi} = \sqrt{(1 - \mu_{GBi}^2) \cdot LSI_i}.
\end{align}
\subsubsection{Ball Compactness and the Scoring Function}
To differentiate stable core balls from uncertain boundary balls, we define ball compactness using the radius as a measure of certainty:
\begin{align}
    \omega_i = \exp\left(-\frac{r_i}{\bar{r} + \epsilon}\right),
\end{align}
where $\bar{r}$ is the average radius of all granular balls. The final scoring function $s_{GBi}$ blends the membership degree with the pythagorean closeness index $\theta_{GBi}$:
\begin{align}
    s_{GBi} = \omega_i \cdot \mu_{GBi} + (1 - \omega_i) \cdot \theta_{GBi}.
\end{align}

\subsection{Fuzzy Granular Ball Random Vector Functional Link Network}
Let $\mathcal{G} = \{(c_1, \delta_{GB_1},y_1),(c_2,\delta_{GB_2},y_2),\ldots,(c_k,\delta_{GB_k},y_k) \},$ where $c_i$ and $\delta_{GB_i}$ denote the center and membership degree of the granular ball $GB_i$, respectively, and $y_i$ is its associated class label.

Let the hidden-layer matrix be denoted by $\mathcal{K}$, which is defined as: 
\begin{align}
\label{eq:66}
    \mathcal{K} = \phi(\mathcal{C}\mathcal{W}+ b) \in \mathbb{R}^{k \times h},
\end{align}
where $\phi$ is the activation function, $\mathcal{W} \in \mathbb{R}^{d \times h}$ is the randomly initialized weight matrix and $b \in \mathbb{R}^{1 \times h}$ denotes the bias vector.

The optimization problem of the proposed GBFRVFL model is given by:
\begin{align}
\label{eq:6}
    \Theta&_{\min}  = \underset{\Theta}{\arg \min} ~ \frac{\mathcal{D}}{2} \|S\eta\|^2 + \frac{1}{2} \|\Theta\|^2 \nonumber \\
    & \quad~~ \text{s.t.} ~ \mathcal{H}\Theta - \mathcal{Y} = \eta,
\end{align}
where $\mathcal{H} = [h(c_1), h(c_2), \ldots, h(c_k)]^T \in \mathbb{R}^{k \times (d+h)}$, $h(c_i) = [c_i~ \varphi(c_i)]$, $S = diag(\delta_{GB_1}, \delta_{GB_2}, \ldots, \delta_{GB_k})$, and $\eta = [\eta_1, \eta_2, \ldots, \eta_k]^T$ is the error term corresponding to $k$ granular balls. The Lagrangian corresponding to the problem \eqref{eq:6} is given by
\begin{align}
    \mathcal{L}(\Theta, \eta, \lambda) = \frac{\mathcal{D}}{2} \|S\eta\|^2 + \frac{1}{2} \|\Theta\|^2 - \lambda^T (\mathcal{H}\Theta - \mathcal{Y} - \eta),
\end{align}
where $\lambda$ is the Lagrangian multiplier. Using the Karush-Kuhn-Tucker (K.K.T.) conditions, we have
\begin{align}
    & \frac{\partial \mathcal{L}}{\partial \Theta} = \Theta - \mathcal{H}^T \lambda, \label{eq:8} \\
    & \frac{\partial \mathcal{L}}{\partial \eta} = \mathcal{D}S^T(S\eta) + \lambda, \label{eq:9} \\
    & \frac{\partial \mathcal{L}}{\partial \lambda} = \mathcal{H}\Theta - \mathcal{Y} - \eta.  \label{eq:10}
\end{align}
By putting Eq. \eqref{eq:10} into Eq. \eqref{eq:9}, we obtain:
\begin{align}
    \lambda = - \mathcal{D}S^TS(\mathcal{H}\Theta - \mathcal{Y}).
\end{align}
By replacing $\lambda$ with the value derived in Eq. \eqref{eq:8}, we get
\begin{align}
    & \Theta = \mathcal{H}^T(- \mathcal{D}S^TS(\mathcal{H}\Theta - \mathcal{Y})), \nonumber \\
    \implies & \Theta = \left(\frac{1}{\mathcal{D}}I + (S\mathcal{H})^T(S\mathcal{H})  \right)^{-1} (S\mathcal{H})^T S \mathcal{Y}, \label{eq:12}
\end{align}
where $I$ is the identity matrix of appropriate dimension. Next, by substituting the expressions from Eqs. \eqref{eq:8}and \eqref{eq:10} into Eq. \eqref{eq:9}, we obtain
\begin{align}
   & \lambda = - \mathcal{D}S^TS(\mathcal{H}\mathcal{H}^T\lambda - \mathcal{Y}), \nonumber \\
   \implies & \lambda + \mathcal{D}S^TS\mathcal{H}\mathcal{H}^T\lambda = \mathcal{D}S^TS \mathcal{Y}, \nonumber \\
   \implies & \lambda = \left(\frac{1}{\mathcal{D}}I + S^TS\mathcal{H}\mathcal{H}^T \right)^{-1}  S^TS \mathcal{Y}.
\end{align}
Put the value of $\lambda$ in \eqref{eq:8}, we get
\begin{align}
\label{eq:14}
    \Theta = \mathcal{H}^T \left(\frac{1}{\mathcal{D}}I + S^TS\mathcal{H}\mathcal{H}^T \right)^{-1}  S^TS \mathcal{Y}.
\end{align}
Two equations can compute $\Theta$, requiring matrix inversion.
If $(d+h) \le k$, Eq. \eqref{eq:12} is used; if $(d+h) > k$, Eq. \eqref{eq:14} is used. This allows inversion in feature or sample space, giving the optimal solution of Eq. \eqref{eq:6}.
\begin{align}
\label{OUTPUT_La}
    \Theta =
\begin{cases}
\left(\frac{1}{\mathcal{D}}I + (S\mathcal{H})^T(S\mathcal{H})  \right)^{-1} (S\mathcal{H})^T S \mathcal{Y}, & \text{if } (d+h) \le k, \\
\mathcal{H}^T \left(\frac{1}{\mathcal{D}}I + S^TS\mathcal{H}\mathcal{H}^T \right)^{-1}  S^TS \mathcal{Y}, & \text{if } k < (d+h).
\end{cases}
\end{align}

The algorithms of the proposed GBFRVFL model are given in Algorithm \ref{Algorithm_GBFRVFL.}.

\begin{algorithm}[ht!]
\caption{Algorithm of the proposed GBFRVFL model.}
\label{Algorithm_GBFRVFL.}
\textbf{Input:} Training set $\mathcal{M}$, and the purity threshold $\rho$. \\
\textbf{Output:} The output weight matrix of the GBFRVFL model. \\ \vspace{-5mm}
\begin{algorithmic}[1]
\State Initialize the full dataset as a granular ball $GB = \mathcal{M}$ and set the granular-ball set $\mathcal{G}$ to empty.
\State $\Psi =\{GB\}$. 
\For{$i = 1:\lvert \Psi \rvert$}
\If{$\mathrm{pur}(GB_i) < \rho$}
    \State Split $GB_i$ into two subsets, $GB_{i1}$ and $GB_{i2}$, using 2-means clustering.
    \State $\Psi \leftarrow \Psi \cup \{GB_{i1}, GB_{i2}\}$
\ElsIf{$pur(GB_i)\geq \rho$}
\State Compute the centroid of $GB_i$ as $c_i = \frac{1}{n_i} \sum_{i=1}^{n_i} x_i$, where $n_i$ is the number of samples in $GB_i$.
\State Assign $GB_i$ the label $y_i$ of its most frequent class.
\State Put $GB_i = \{(c_i,y_i)\}$ in $\mathcal{G}$. 
\EndIf 
\EndFor 
\If{$\Psi \neq \emptyset$}
    \State Return to Step 3 to continue hierarchical decomposition.
\Else
    \State Compute the membership degree of $GB_i$ as $\delta_{GB_i}$.
\EndIf
\State Form the set $\mathcal{G} = {GB_i}_{i=1}^k = {(c_i, \delta_{GB_i}, y_i)}_{i=1}^k$, with each ball defined by its center, membership degree, and label.
\State Compute the hidden-layer features using \eqref{eq:66}.
\State Calculate the enhanced features matrix using $\mathcal{H} = [\mathcal{C}, ~ \mathcal{K}]$. 
\State Calculate the output layer weights using \eqref{OUTPUT_La}.
\end{algorithmic}
\end{algorithm}

\section{Numerical Experiments and Results}
To evaluate the GBFRVFL framework, two membership strategies are integrated into the GBRVFL architecture: \textbf{F-GBRVFL} (fuzzy membership) and \textbf{SDAP-GBRVFL} (SDAP membership). Both models are tested on 37 UCI and KEEL benchmark datasets and compared with RVFL \cite{pao1994learning}, ELM \cite{huang2006extreme}, GB-RVFL and GE-GB-RVFL \cite{sajid2025gb}, and CRVFL and ACRVFL \cite{liu2025complex}. Experiments with added label noise are provided in Section S.I, and sensitivity analyses are presented in Section S.II of the supplementary material.

\subsection{Experimental Setup}
Experiments are run on a Windows 11 workstation with an Intel Xeon Gold 6226R (2.90 GHz) and 256GB RAM using Python 3.11. Datasets are split 70:30 for training and testing, with hyperparameters tuned via grid search and five-fold cross-validation, where the regularization parameters are explored within the set $\mathcal{D} = \{10^{-5}, 10^{-4}, \ldots, 10^{5}\}$. Hidden nodes  $h_l$ range from $3$ to $203$, and nine activation functions are tested: SELU, ReLU, Sigmoid, Sine, Hardlim, Tribas, Radbas, Sign, and Leaky ReLU.

\begin{table*}[ht!]
\centering
    \caption{Performance comparison of the proposed F-GBRVFL and SDAP-GBRVFL models along with the baseline models for UCI and KEEL datasets.}
    \label{Classification performance UCI and KEEL}
    \resizebox{1\linewidth}{!}{
\begin{tabular}{lcccccccc} 
\hline
Dataset & RVFL \cite{pao1994learning} & ELM \cite{huang2006extreme} & GB-RVFL \cite{sajid2025gb} & GE-GB-RVFL \cite{sajid2025gb} & CRVFL \cite{liu2025complex} & ACRVFL \cite{liu2025complex} & F-GBRVFL$^{\dagger}$ & SDAP-GBRVFL$^{\dagger}$ \\
\hline
bank & 89.17 & 87.31 & 88.43 & 88.95 & 88.28 & 88.8 & 89.09 & 89.24 \\
blood & 76.44 & 72.14 & 76 & 77.33 & 75.56 & 77.33 & 79.11 & 80.33 \\
breast\_cancer & 72.09 & 74.42 & 77.91 & 65.12 & 79.07 & 74.42 & 74.42 & 74.93 \\
breast\_cancer\_wisc\_prog & 75 & 73.33 & 73.33 & 73.33 & 66.67 & 66.67 & 76.67 & 76.67 \\
bupa or liver-disorders.csv & 65.38 & 65.38 & 54.81 & 56.73 & 94.95 & 77.24 & 62.5 & 67.54 \\
checkerboard\_Data.csv & 85.94 & 85.98 & 87.02 & 87.02 & 62.48 & 71.5 & 87.02 & 87.5 \\
chess\_krvkp & 90.41 & 90.2 & 91.45 & 89.16 & 93.95 & 94.06 & 93.64 & 94.58 \\
cleve.csv & 81.11 & 80 & 75.56 & 82.22 & 73.79 & 79.79 & 81.11 & 81.11 \\
conn\_bench\_sonar\_mines\_rocks & 74.6 & 71.43 & 74.6 & 68.25 & 65.08 & 66.67 & 73.9 & 74.43 \\
credit\_approval & 84.62 & 82.3 & 77.88 & 80.77 & 83.65 & 84.13 & 84.65 & 84.62 \\
crossplane150.csv & 81.11 & 81.11 & 86.67 & 73.33 & 84.68 & 73.79 & 83.33 & 88.89 \\
cylinder\_bands & 72.73 & 70.67 & 59.74 & 64.29 & 70.78 & 66.23 & 71.43 & 72.88 \\
ecoli-0-1-4-6\_vs\_5.csv & 98.81 & 98.81 & 98.81 & 98.81 & 87.14 & 86.74 & 98.81 & 99.62 \\
ecoli-0-1-4-7\_vs\_5-6.csv & 90 & 96 & 94 & 94 & 72.9 & 70.66 & 70 & 96 \\
fertility & 90 & 80 & 86.67 & 90 & 90 & 90 & 90 & 90 \\
haber.csv & 76.09 & 78.26 & 78.26 & 77.17 & 75.46 & 85.85 & 79.35 & 80.17 \\
haberman.csv & 76.09 & 76.09 & 78.26 & 77.17 & 71.58 & 83.33 & 83.35 & 84.17 \\
haberman\_survival & 76.09 & 78.26 & 78.26 & 77.17 & 82.61 & 82.61 & 82.35 & 83.26 \\
heart\_hungarian & 72.78 & 72.65 & 74.16 & 74.16 & 75.28 & 76.4 & 75.28 & 80.9 \\
hepatitis & 72.34 & 70.11 & 72.34 & 72.34 & 76.6 & 82.98 & 72.34 & 76.6 \\
hill\_valley & 68.41 & 67.31 & 59.07 & 60.71 & 53.3 & 54.12 & 66.48 & 68.13 \\
horse\_colic & 83.78 & 80.18 & 72.97 & 76.58 & 76.58 & 77.48 & 83.78 & 86.49 \\
ionosphere & 82.79 & 82.45 & 83.96 & 83.02 & 85.85 & 83.96 & 84.91 & 83.96 \\
monks\_3 & 90.41 & 90.41 & 85.63 & 91.02 & 74.85 & 85.03 & 92.22 & 90.42 \\
new-thyroid1.csv & 86 & 86 & 90.77 & 96.92 & 73.15 & 67.62 & 89.23 & 100 \\
oocytes\_merluccius\_nucleus\_4d & 80.71 & 79.74 & 80.78 & 78.5 & 67.43 & 77.2 & 81.76 & 81.11 \\
oocytes\_trisopterus\_nucleus\_2f & 80.12 & 80.94 & 75.55 & 80.29 & 69.71 & 71.9 & 83.21 & 76.28 \\
statlog\_australian\_credit & 68.75 & 68.75 & 62.02 & 62.02 & 70.19 & 70.19 & 70.67 & 71.63 \\
statlog\_german\_credit & 70.33 & 70.67 & 70 & 75.33 & 69 & 66.33 & 77 & 75.33 \\
vehicle2.csv & 90.85 & 90.03 & 91.73 & 92.52 & 90.36 & 85.9 & 94.88 & 97.24 \\
vertebral\_column\_2clases & 81.4 & 72.25 & 74.19 & 74.19 & 66.67 & 77.42 & 88.17 & 87.1 \\
votes.csv & 90.47 & 90.47 & 92.37 & 87.02 & 67.22 & 67.78 & 95.42 & 97.71 \\
vowel.csv & 99.33 & 98.99 & 85.19 & 63.97 & 70.71 & 67.03 & 94.95 & 92.59 \\
wpbc.csv & 69.49 & 70.97 & 76.27 & 69.49 & 75.69 & 84.96 & 64.41 & 72.88 \\
yeast-0-2-5-6\_vs\_3-7-8-9.csv & 89.71 & 89.71 & 93.38 & 90.73 & 77.16 & 67.02 & 93.71 & 92.72 \\
yeast-0-2-5-7-9\_vs\_3-6-8.csv & 95.35 & 95.68 & 98.01 & 97.35 & 78.44 & 70.71 & 88.74 & 97.02 \\
yeast-0-3-5-9\_vs\_7-8.csv & 88.82 & 90.79 & 69.74 & 45.39 & 70.01 & 81.01 & 86.84 & 90.79 \\ \hline
Average Acc & 81.55 & 80.81 & 79.62 & 78.17 & 75.86 & 76.62 & \underline{82.29} & \textbf{84.46} \\ \hline
Average Rank & 4.51 & 5.22 & 4.86 & 5.16 & 5.72 & 5.26 & 3.2 & 2.07 \\
\hline
\multicolumn{9}{l}{The proposed model is denoted by $^{\dagger}$.}\\
 \multicolumn{9}{l}{The top and second-best models in terms of ACC are denoted by boldface and underline, respectively.}
\end{tabular}
}
\end{table*}

\subsection{Evaluation on UCI and KEEL Datasets}
The classification effectiveness of the proposed F-GBRVFL and SDAP-GBRVFL models, along with the corresponding results of the baseline models, is evaluated in terms of classification accuracy (Acc), and the comparative outcomes are given in Table \ref{Classification performance UCI and KEEL}. The proposed  F-GBRVFL and SDAP-GBRVFL models achieve an average Acc of $82.29\%$ and $84.46$, whereas the baseline models RVFL, ELM, GB-RVFL, GE-GB-RVFL, CRVFL, and ACRVFL attain an average Acc of $81.55\%$, $80.81\%$, $79.62\%$, $78.17\%$, $75.86\%$, and  $76.62\%$, respectively. In terms of average Acc, the proposed F-GBRVFL and SDAP-GBRVFL models consistently outperform the competing models, demonstrating their robust and dependable predictive performance. However, while average Acc offers an overall performance, it may not fully reflect model behavior, since superior results on some datasets can be offset by relatively weaker performance on others. To address this limitation and provide a more rigorous assessment of whether the observed performance differences are statistically meaningful, a series of nonparametric statistical tests, as recommended by \citet{demvsar2006statistical}, are employed. These statistical methods are suitable for comparing multiple learning algorithms across diverse datasets when parametric test assumptions are not met. Accordingly, this study uses nonparametric techniques, including rank-based methods, the Friedman test, and Nemenyi post hoc analysis, to ensure objective performance comparisons. In rank-based evaluation, models are ranked within each dataset and aggregated across datasets, with better-performing models receiving lower ranks and poorer-performing models receiving higher ranks. For a comparison involving $\xi$ models evaluated on $N$ datasets, let $\mathcal{R}_i^j$ denote the rank assigned to the $j^{th}$ model on the $i^{th}$ dataset. The overall average rank of the $j^{th}$ model is then calculated as $\mathcal{R}^j = \frac{1}{n} \sum_{i=1}^{N} \mathcal{R}_i^j.$ The average rank of our proposed F-GBRVFL and SDAP-GBRVFL models, along with the baseline RVFL, ELM, GB-RVFL, GE-GB-RVFL, CRVFL, and ACRVFL models, are $3.2$, $2.07$, $4.51$, $5.22$, $4.86$, $5.16$, and $5.26$, respectively. The F-GBRVFL and SDAP-GBRVFL models attain the smallest average rank across all evaluated models. As lower rank values indicate better performance, this result highlights the proposed F-GBRVFL and SDAP-GBRVFL as the most effective approaches among the models under comparison. The Friedman test \cite{friedman1937use} is a nonparametric statistical technique employed to examine whether meaningful performance differences exist among multiple models by comparing their average ranks over several datasets. The null hypothesis assumes no significant differences in model performance. The Friedman test statistic, denoted by $\chi_F^2$, is assumed to follow a chi-square distribution with $(\xi - 1)$ degrees of freedom and is calculated as $\chi_F^2 = \frac{12N}{\xi(\xi+1)} \left[ \sum_{j} \mathcal{R}_j^2 - \frac{\xi(\xi+1)^2}{4} \right].$ In addition, the corresponding $F_F$ statistic is given by $F_F = \frac{(N - 1)\chi_F^2}{N(\xi - 1) - \chi_F^2},$ which follows an $F$-distribution with $(\xi - 1)$ and $(N - 1)(\xi - 1)$ degrees of freedom. For $N = 37$ datasets and $\xi = 8$ competing models, the Friedman test produces a test statistic of $\chi_F^2 = 66.26$, leading to an $F_F$ value of $12.38$. At the $5\%$ significance level, the critical value of the $F$-distribution with  $(7,252)$ degrees of freedom is $2.05$. Since $F_F$ exceeds the threshold, the null hypothesis is rejected, indicating significant performance differences. Next, the Nemenyi post hoc test is employed to analyze the pairwise differences in performance between the models. The critical difference (C.D.) is determined using: $\text{C.D.} = q_\alpha \sqrt{\frac{\xi(\xi+1)}{6N}},$ where $q_\alpha$ represents the critical value from the two-tailed Nemenyi distribution. At a $5\%$ significance level, $q_\alpha = 3.031$, which yields a corresponding C.D. of $1.72$. The differences in average ranks between the proposed F-GBRVFL and SDAP-GBRVFL models and the baseline RVFL, ELM, GB-RVFL, GE-GB-RVFL, CRVFL, and ACRVFL models are $(1.31, 2.44)$, $(2.02, 3.15)$, $(1.66, 2.79)$, $(1.96, 3.09)$, $(2.52, 3.65)$, and $(2.06, 3.19)$, respectively. This difference indicates that F-GBRVFL outperforms all baseline models except RVFL and GB-RVFL, showing consistently strong performance, while SDAP-GBRVFL achieves statistically significant improvements over the baselines.

\section{Conclusion}
In this paper, we introduced a novel granular-ball-based randomized learning framework, termed GBFRVFL, which leverages the coarse-grained abstraction of granular balls to enhance robustness and generalization in the presence of noise, outliers, and class imbalance. Two variants of the proposed model are developed to address uncertainty and improve classification performance. The first, F-GBRVFL, integrates a fuzzy membership scheme to assign degrees of reliability to each granular ball, effectively mitigating the influence of noisy or ambiguous samples. The second, SDAP-GBRVFL, incorporates a novel statistical density-adaptive pythagorean membership (SDAPM) scheme that dynamically adjusts membership and non-membership values based on class variance, local sparsity, and granular-ball compactness, ensuring effective handling of uncertain granular-ball distributions and class imbalance. Extensive experiments on UCI and KEEL benchmark datasets, including scenarios with varying levels of label noise, demonstrate that both F-GBRVFL and SDAP-GBRVFL consistently outperform existing randomized learning and granular ball-based models. SDAP-GBRVFL shows strong robustness to noise and better preserves intrinsic data structure while retaining RVFL’s computational efficiency. Statistical analyses confirm significant improvements, validating the proposed membership schemes. This work focuses on shallow RVFL networks, with future extensions to deep and ensemble models.

\bibliographystyle{IEEEtranN}
\bibliography{refs.bib}

\clearpage
\section*{Supplementary Material}

\renewcommand{\thesection}{S.I}
\section{Results and Discussion on UCI and KEEL Datasets in Noisy Environments}
The UCI and KEEL datasets considered in this subsection represent realistic learning scenarios in which training data are frequently corrupted by label noise. Evaluating model robustness under such adverse conditions is therefore essential for assessing the reliability and practical applicability of learning algorithms. To this end, controlled levels of label noise are intentionally introduced into five representative datasets, namely cleve, conn\_bench\_sonar\_mines\_rocks, ecoli-0-1-4-6\_vs\_5, haberman\_survival, and ionosphere, with noise ratios set to 5\%, 10\%, 20\%, 30\%, and 40\%. The performance of the proposed F-GBRVFL and SDAP-GBRVFL models is compared against the baseline models. The corresponding classification Acc are reported in Table \ref{Classification label noise}. From the results, it is evident that both proposed models demonstrate strong resilience to increasing levels of label noise across all datasets. In particular, the SDAP-GBRVFL model consistently achieves the highest or second-highest Acc in most noise settings, highlighting the effectiveness of the SDAP membership mechanism in suppressing the influence of mislabeled samples. On the cleve dataset, SDAP-GBRVFL attains the best average Acc among all competing models, maintaining stable performance even when the noise level reaches 40\%. This indicates that the adaptive membership weighting successfully mitigates the adverse impact of label corruption. Similar trends can be observed on the conn\_bench\_sonar\_mines\_rocks dataset, where both F-GBRVFL and SDAP-GBRVFL outperform conventional RVFL-based models across nearly all noise levels. On the ecoli-0-1-4-6\_vs\_5 dataset, the proposed models consistently deliver strong performance, with SDAP-GBRVFL achieving the best overall average Acc. This behavior confirms that granular-ball abstraction, combined with adaptive fuzzy membership assignment, effectively reduces the influence of corrupted labels. On the ionosphere dataset, both proposed models maintain stable Acc as noise increases, outperforming most baseline models at higher noise levels and demonstrating improved generalization under noisy conditions. When considering the overall average Acc across all datasets and noise ratios, SDAP-GBRVFL achieves the best performance, followed closely by F-GBRVFL. These results clearly indicate that incorporating fuzzy and statistical density-adaptive Pythagorean memberships into the RVFL framework substantially enhances robustness to label noise. 

\renewcommand{\thetable}{S.I}
\begin{table*}[ht!]
\centering
    \caption{Performance comparison of the proposed F-GBRVFL and SDAP-GBRVFL models along with the baseline models for UCI and KEEL datasets with contaminated label noise.}
    \label{Classification label noise}
    \resizebox{1\linewidth}{!}{
\begin{tabular}{lccccccccc}
 \hline
Dataset & Noise & RVFL \cite{pao1994learning} & ELM \cite{huang2006extreme} & GB-RVFL \cite{sajid2025gb} & GE-GB-RVFL \cite{sajid2025gb} & CRVFL \cite{liu2025complex} & ACRVFL \cite{liu2025complex} & F-GBRVFL$^{\dagger}$ & SDAP-GBRVFL$^{\dagger}$ \\
\hline
cleve & 5\% & 75.11 & 70.78 & 70 & 73.33 & 70.56 & 70.56 & 78.89 & 80 \\
 & 10\% & 82.22 & 80 & 81.11 & 61.11 & 65.56 & 65.56 & 82.22 & 81.11 \\
 & 20\% & 80 & 75.11 & 75.56 & 66.67 & 65.56 & 64.89 & 76.67 & 77.78 \\
 & 30\% & 72.22 & 75.56 & 63.33 & 57.78 & 62.36 & 62.36 & 76.67 & 74.44 \\
 & 40\% & 57.78 & 58.89 & 66.67 & 52.22 & 45.56 & 45.56 & 53.33 & 57.78 \\ \hline
Average Acc &  & 73.47 & 72.07 & 71.33 & 62.22 & 61.92 & 61.79 & 73.56 & 74.22 \\ \hline
conn\_bench\_sonar\_mines\_rocks & 5\% & 76.19 & 63.49 & 71.43 & 76.19 & 66.67 & 68.25 & 76.19 & 79.37 \\
 & 10\% & 70.19 & 70.78 & 69.84 & 73.02 & 68.25 & 65.08 & 78.38 & 80.95 \\
 & 20\% & 70.95 & 76.19 & 74.6 & 77.78 & 69.84 & 69.84 & 74.6 & 72.25 \\
 & 30\% & 71.43 & 73.02 & 71.43 & 76.19 & 52.38 & 52.38 & 72.21 & 73.73 \\
 & 40\% & 66.67 & 52.38 & 69.84 & 69.84 & 41.27 & 42.86 & 70.44 & 71.79 \\ \hline
Average Acc &  & 71.09 & 67.17 & 71.43 & 74.6 & 59.68 & 59.68 & 74.36 & 75.62 \\ \hline
ecoli-0-1-4-6\_vs\_5 & 5\% & 90 & 90 & 83.33 & 94.05 & 90.95 & 90.95 & 96.43 & 98.81 \\
 & 10\% & 98.81 & 98.81 & 98.81 & 94.05 & 95.95 & 95.95 & 95.24 & 94.05 \\
 & 20\% & 98.81 & 98.81 & 98.81 & 100 & 95.95 & 95.95 & 98.67 & 98.05 \\
 & 30\% & 90.86 & 90.05 & 94.05 & 96.43 & 55.95 & 55.95 & 91.67 & 94.05 \\
 & 40\% & 64.05 & 61.67 & 65.48 & 58.33 & 45.95 & 45.95 & 64.76 & 69.52 \\ \hline
Average Acc &  & 88.51 & 87.87 & 88.1 & 88.57 & 76.95 & 76.95 & 89.35 & 90.9 \\ \hline
haberman\_survival & 5\% & 72.17 & 72.09 & 75 & 77.17 & 83.7 & 82.61 & 67.39 & 80.43 \\
 & 10\% & 77.17 & 75.35 & 76.09 & 77.17 & 79.35 & 73.91 & 81.52 & 85.22 \\
 & 20\% & 79.35 & 77.17 & 75.35 & 77.17 & 75 & 82.61 & 81.52 & 80.43 \\
 & 30\% & 51.09 & 59.35 & 76.09 & 61.96 & 82.61 & 80.43 & 80.74 & 81.52 \\
 & 40\% & 55.43 & 52.17 & 56.52 & 53.91 & 53.26 & 58.7 & 60.43 & 62.91 \\ \hline
Average Acc &  & 67.04 & 67.23 & 71.81 & 69.48 & 74.78 & 75.65 & 74.32 & 78.1 \\ \hline
ionosphere & 5\% & 85.68 & 85.74 & 86.79 & 83.02 & 83.02 & 85.85 & 86.79 & 86.79 \\
 & 10\% & 86.74 & 82.08 & 86.79 & 87.74 & 83.96 & 83.96 & 88.68 & 86.79 \\
 & 20\% & 85.58 & 83.96 & 82.74 & 81.13 & 83.02 & 82.08 & 83.96 & 83.02 \\
 & 30\% & 83.96 & 77.36 & 70.75 & 74.34 & 77.92 & 73.58 & 78.49 & 78.87 \\
 & 40\% & 55.6 & 53.77 & 58.49 & 55.66 & 47.55 & 40.81 & 59.89 & 63.21 \\ \hline
Average Acc &  & 79.51 & 76.58 & 77.11 & 76.38 & 75.09 & 73.26 & 79.56 & 79.74 \\ \hline
Overall Average Acc &  & 75.92 & 74.18 & 75.96 & 74.25 & 69.68 & 69.47 & \underline{78.23} & \textbf{79.72} \\
\hline
\multicolumn{9}{l}{The proposed model is denoted by $^{\dagger}$.}\\
 \multicolumn{9}{l}{The top and second-best models in terms of ACC are denoted by boldface and underline, respectively.}
\end{tabular}
}
\end{table*}

\renewcommand{\thesection}{S.II}
\section{Sensitivity Analysis}
This subsection examines the robustness and stability of the proposed F-GBRVFL and SDAP-GBRVFL models by analyzing their sensitivity to key hyperparameters. The study evaluates how changes in the regularization parameter, choice of activation function, and model capacity affect classification performance across multiple datasets.

\renewcommand{\thefigure}{S.1}
\begin{figure*}[ht!]
\begin{minipage}{.246\linewidth}
\centering
\subfloat[cleve (F-GBRVFL)]{\label{1a}\includegraphics[scale=0.24]{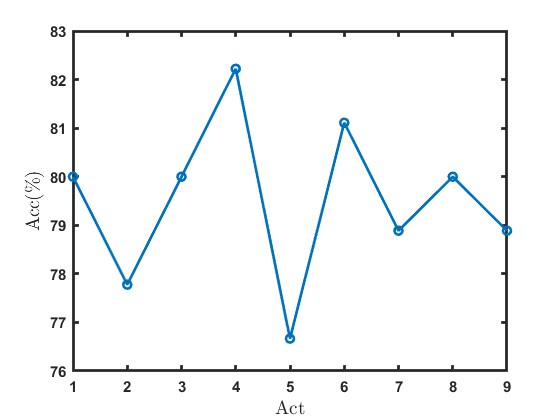}}
\end{minipage}
% \par\medskip
% \par\medskip
\begin{minipage}{.246\linewidth}
\centering
\subfloat[ecoli-0-1-4-6\_vs\_5 (F-GBRVFL)]{\label{1b}\includegraphics[scale=0.24]{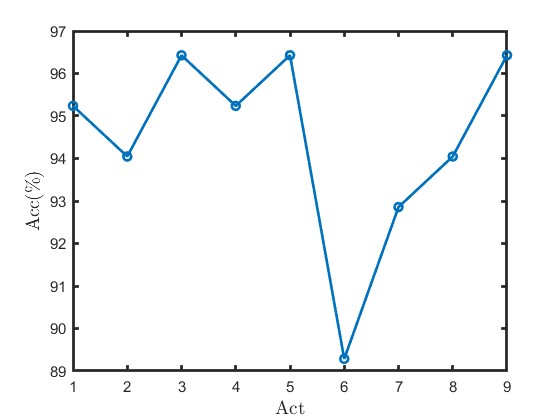}}
\end{minipage}
% \par\medskip
\begin{minipage}{.246\linewidth}
\centering
\subfloat[cleve (SDAP-GBRVFL)]{\label{1c}\includegraphics[scale=0.24]{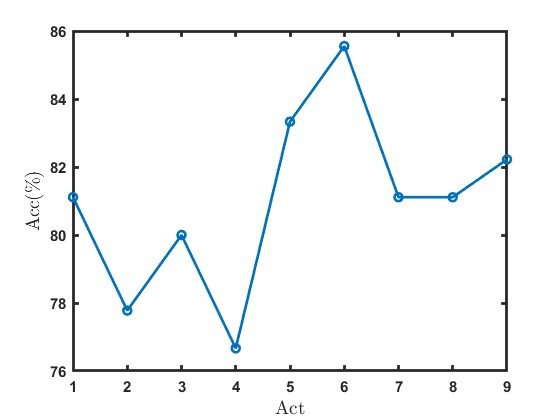}}
\end{minipage}
% \par\medskip
% \par\medskip
\begin{minipage}{.246\linewidth}
\centering
\subfloat[ecoli-0-1-4-6\_vs\_5 (SDAP-GBRVFL)]{\label{1d}\includegraphics[scale=0.24]{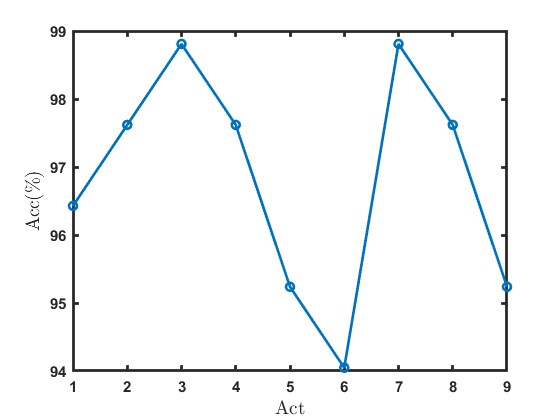}}
\end{minipage}
\caption{The impact of changing the Act on the Acc values of the proposed F-GBRVFL and SDAP-GBRVFL models.}
\label{Effect of parameters Act}
\end{figure*}

\renewcommand{\thefigure}{S.2}
\begin{figure*}[ht!]
\begin{minipage}{.246\linewidth}
\centering
\subfloat[cleve (F-GBRVFL)]{\label{2a}\includegraphics[scale=0.24]{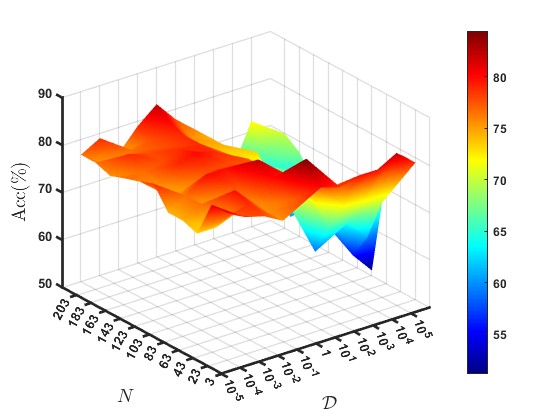}}
\end{minipage}
% \par\medskip
% \par\medskip
\begin{minipage}{.246\linewidth}
\centering
\subfloat[ecoli-0-1-4-6\_vs\_5 (F-GBRVFL)]{\label{2b}\includegraphics[scale=0.24]{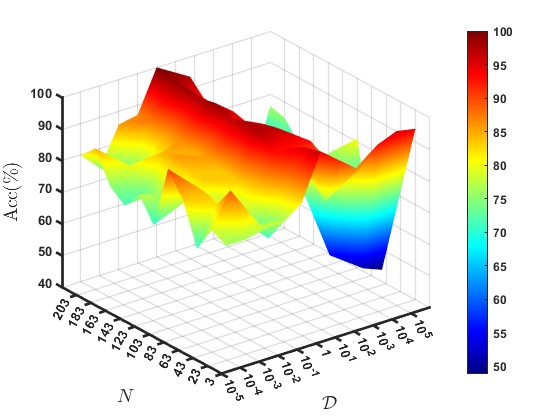}}
\end{minipage}
\begin{minipage}{.246\linewidth}
\centering
\subfloat[cleve (SDAP-GBRVFL)]{\label{2c}\includegraphics[scale=0.24]{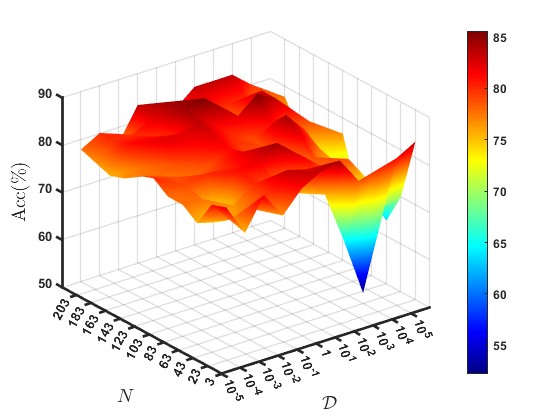}}
\end{minipage}
% \par\medskip
% \par\medskip
% \par\medskip
\begin{minipage}{.246\linewidth}
\centering
\subfloat[ecoli-0-1-4-6\_vs\_5 (SDAP-GBRVFL)]{\label{2d}\includegraphics[scale=0.24]{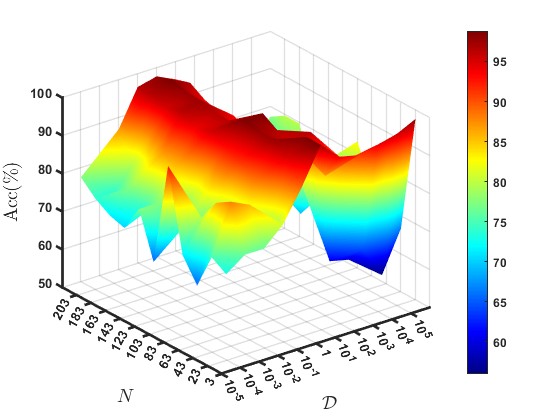}}
\end{minipage}
\caption{The impact of changing the parameters $\mathcal{D}$ and $N$ on the Acc values of the proposed F-GBRVFL and SDAP-GBRVFL models.}
\label{Effect of parameters D and N}
\end{figure*}

\subsection{Sensitivity analysis of the activation function \texorpdfstring{$\text{(Act)}$}{Act} }
Figure~\ref{Effect of parameters Act} illustrates the influence of different activation functions on the classification Acc of the proposed F-GBRVFL and SDAP-GBRVFL models across two representative datasets, namely cleve and ecoli-0-1-4-6\_vs\_5. Nine commonly used activation functions are evaluated to analyze the robustness of the proposed models with respect to activation choice. As shown in Fig.~\ref{1a} and Fig.~\ref{1b}, F-GBRVFL exhibits relatively stable performance across most activation functions, with peak Acc achieved using mid-range nonlinear activations, indicating that the fuzzy membership mechanism effectively alleviates sensitivity to activation selection. However, certain activation functions lead to noticeable performance drops, particularly in more complex or imbalanced datasets, reflecting the inherent variability of random hidden mappings. In contrast, Fig.~\ref{1c} and Fig.~\ref{1d} demonstrate that SDAP-GBRVFL consistently attains higher Acc and exhibits reduced performance fluctuation across different activation functions. This improved stability can be attributed to the proposed SDAPM scheme, which adaptively adjusts membership and non-membership values based on statistical density, local sparsity, and granular-ball compactness. 

\subsection{Sensitivity analysis of hyperparameters \texorpdfstring{$\mathcal{D}$}{D} and \texorpdfstring{$N$}{N} }
Figure~\ref{Effect of parameters D and N} illustrates the joint sensitivity analysis of the hyperparameters $\mathcal{D}$ and $N$, which respectively control the regularization strength and the number of hidden nodes in the proposed F-GBRVFL and SDAP-GBRVFL models. The parameters are varied over a wide range to assess model robustness and stability under different configurations. For the cleve dataset using F-GBRVFL (Fig.~\ref{2a}), high and stable Acc values are observed when $\mathcal{D}$ assumes moderate values and $N$ lies in an intermediate range. In contrast, extreme settings of either parameter, particularly very small $\mathcal{D}$ or excessively large $N$, lead to noticeable performance degradation, indicating overfitting or insufficient regularization. For the ecoli-0-1-4-6\_vs\_5 dataset under the F-GBRVFL framework (Fig.~\ref{2b}), the model demonstrates strong sensitivity to $\mathcal{D}$, where moderate-to-large values consistently yield superior Acc across a broad range of $N$. In the case of SDAP-GBRVFL on the cleve dataset (Fig.~\ref{2c}), the Acc surface appears smoother and more stable, with peak performance attained across a wider region of the parameter space. Similarly, for ecoli-0-1-4-6\_vs\_5 using SDAP-GBRVFL (Fig.~\ref{2d}), the model achieves consistently high Acc over a broad range of $\mathcal{D}$ and $N$ values, with only marginal degradation at extreme configurations.

\end{document}